\documentclass[10pt,twocolumn,letterpaper]{article}

\usepackage{iccv}
\usepackage{times}
\usepackage{epsfig}
\usepackage{graphicx}
\usepackage{amsmath}
\usepackage{amssymb}
\usepackage{epsfig}
\usepackage{graphicx}
\usepackage{amsmath}
\usepackage{amssymb}
\usepackage{mathabx}
\usepackage{stmaryrd}
\usepackage{wasysym}
\usepackage{algorithm}
\usepackage{algorithmic}
\usepackage{graphicx}
\usepackage{textcomp}
\usepackage{tikz}
\usepackage{pgfpages}
\usepackage{pgfplots}
\usepackage{amsmath}
\usepackage{subfig}

\usepackage[pagebackref=true,breaklinks=true,letterpaper=true,colorlinks,bookmarks=false]{hyperref}

\usepackage[breaklinks=true,bookmarks=false]{hyperref}

\iccvfinalcopy 

\def\iccvPaperID{****} 

\ificcvfinal\fi

\begin{document}

\title{DNN Quantization with Attention}

\author{Ghouthi Boukli Hacene\\
MILA, IMT Atlantique\\

{\tt\small ghouthi.bouklihacene@imt-atlantique.fr}
\and
Lukas Mauch\\
Sony Europe B.V., Stuttgart, Germany\\
{\tt\small lukas.mauch@sony.com}
\and
Stefan Uhlich\\
Sony Europe B.V., Stuttgart, Germany\\
{\tt\small stefan.uhlich@sony.com}
\and
Fabien Cardinaux\\
Sony Europe B.V., Stuttgart, Germany\\
{\tt\small fabien.cardinaux@sony.com}
}

\maketitle
\ificcvfinal\thispagestyle{empty}\fi

\begin{abstract}
   Low-bit quantization of network weights and activations can drastically reduce the memory footprint, complexity, energy consumption and latency of Deep Neural Networks (DNNs). However, low-bit quantization can also cause a considerable drop in accuracy, in particular when we apply it to complex learning tasks or lightweight DNN architectures. In this paper, we propose a training procedure that relaxes the low-bit quantization. We call this procedure \textit{DNN Quantization with Attention} (DQA). The relaxation is achieved by using a learnable linear combination of high, medium and low-bit quantizations. Our learning procedure converges step by step to a low-bit quantization using an attention mechanism with temperature scheduling. In experiments, our approach outperforms other low-bit quantization techniques on various object recognition benchmarks such as CIFAR10, CIFAR100 and ImageNet ILSVRC 2012, achieves almost the same accuracy as a full precision DNN, and considerably reduces the accuracy drop when quantizing lightweight DNN architectures.
\end{abstract}

\section{Introduction}
During the last decade, Deep Neural Networks (DNNs) in general and Convolutional Neural Networks (CNNs)~\cite{lecun1998gradient} in particular became state-of-the-art in many computer vision tasks, such as image classification, object detection/segmentation, and face recognition~\cite{iandola2016squeezenet,DBLP:journals/corr/SimonyanZ14a,DBLP:journals/corr/Graham14a,szegedy2015rethinking}. However, to be state-of-the-art, DNNs require a large number of trainable parameters, and considerable computational power. Because of such requirements, DNNs are not applicable to embedded systems with few resources or for mobile applications.

Therefore, a large number of different methods have been introduced to alleviate these resource constraints. Among them, identifying the less important DNNs parameters when computing an output, and pruning them in order to reduce both memory footprint and computational power. Several pruning methods have been introduced~\cite{yamamoto2018pcas,ramakrishnan2020differentiable,he2020learning} that can remove DNN parameters, intermediate inputs, and even layers if they are irrelevant for good network performance.

Another way to reduce the complexity of DNNs is quantization. Indeed, it has been shown that low precision networks can be trained even in extreme cases, when using very low bitwidths to encode the weights and activations. Binary networks, for example, constrain the network weights to be either 1 or -1~\cite{courbariaux2015binaryconnect,hubara2016binarized}. However, binarization can cause a big drop in accuracy when applied to complex learning tasks or already compact network architectures. Similar quantization approaches proposed to limit weights to three or more values~\cite{li2016ternary,li2016approximate,zhu2016trained,choi2018bridging,cardinaux2020iteratively}. Compared to binary networks, they have significantly higher performance, but also require more memory and hardware resources. 
Binary-Relax (BR)~\cite{yin2018binaryrelax} proposed to improve the binary and ternary network performance by relaxing the quantization process. Using a linear interpolation between quantized and full precision training, they progressively 
drive the weights of a DNN towards quantized values.
However, the scaling factors for this linear interpolation are handcrafted and predetermined before training. This represents an additional constraint that may prevent the network to converge to the best solution. Moreover, considering more than one quantization scheme during training could help the DNN parameters to converge to a better quantized state. In this procedure, BR becomes more complex, since we have to handcraft a way to efficiently initialize and adapt a bigger set of scaling factors. 

Another line of works showed that the performance of DNNs can be improved if we give them the ability to adapt their own network architecture~\cite{elsken2019neural}. We can, for example, learn the number of bits that are used to quantize the weights in each layer~\cite{uhlich2019differentiable} or the criterion that is  used for pruning network weights or activations~\cite{he2020learning}.

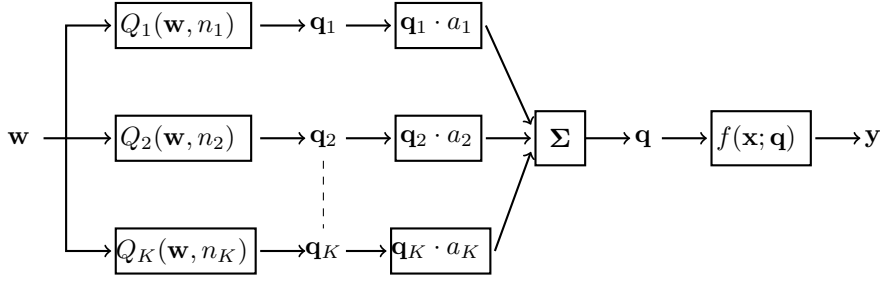
\begin{figure*}
    \centering
    \def \y1{2.5}
    \begin{tikzpicture}[scale=0.6]
    \node[text width=3cm] at (-5,-5) 
         {};
    \node[text width=3cm] at (0,0) 
         {$\mathbf{w}$};
         \draw[thick] (-1.7,0)--(-1.2,0.0);
         \draw[->, thick](-1.2,0.0)--(-1.2,\y1)--(-0.2,\y1);
         \draw[thick](-0.1,\y1+0.5)--(2.9,\y1+0.5)--(2.9,\y1-0.5)--(-0.1,\y1-0.5)--cycle;
         \node[text width=3cm] at (2.5,\y1) 
         {$Q_1(\mathbf{w},n_1)$};
         \draw[->,thick](3.1,\y1)--(4.1,\y1);
         \node[text width=3cm] at (6.7,\y1) 
         {$\mathbf{q}_1$};
        \draw[->,thick](5,\y1)--(6,\y1);
        \draw[thick](6.1,\y1+0.5)--(8,\y1+0.5)--(8,\y1-0.5)--(6.1,\y1-0.5)--cycle;
        \node[text width=3cm] at (8.7,\y1) 
         {$\mathbf{q}_1 \cdot a_1$};
         \draw[->,thick](8.1,\y1)--(9.1,0.3);
        
         \draw[->, thick](-1.2,0.0)--(-0.2,0);
         \draw[thick](-0.1,0+0.5)--(2.9,0+0.5)--(2.9,0-0.5)--(-0.1,0-0.5)--cycle;
         \node[text width=3cm] at (2.5,0) 
         {$Q_2(\mathbf{w}, n_2)$};
         \draw[->,thick](3.1,0)--(4.1,0);
         \node[text width=3cm] at (6.7,0) 
         {$\mathbf{q}_2$};
        \draw[->,thick](5,0)--(6,0);
        \draw[thick](6.1,0+0.5)--(8,0+0.5)--(8,0-0.5)--(6.1,0-0.5)--cycle;
        \node[text width=3cm] at (8.7,0) 
         {$\mathbf{q}_2 \cdot a_2$};
         \draw[->,thick](8.1,0)--(9.1,0);

         \draw[->, thick](-1.2,0.0)--(-1.2,-\y1)--(-0.2,-\y1);
         \draw[thick](-0.1,-\y1+0.5)--(3,-\y1+0.5)--(3,-\y1-0.5)--(-0.1,-\y1-0.5)--cycle;
         \node[text width=3cm] at (2.4,-\y1) 
         {$Q_K(\mathbf{w},n_K)$};
         \draw[->,thick](3.1,-\y1)--(4.1,-\y1);
         \node[text width=3cm] at (6.6,-\y1) 
         {$\mathbf{q}_K$};
        \draw[->,thick](5,-\y1)--(5.9,-\y1);
        \draw[thick](6,-\y1+0.5)--(8.2,-\y1+0.5)--(8.2,-\y1-0.5)--(6,-\y1-0.5)--cycle;
        \node[text width=3cm] at (8.5,-\y1) 
         {$\mathbf{q}_K \cdot a_K$};
         \draw[->,thick](8.3,-\y1)--(9.1,-0.3);
         
         \draw[thick](9.2,0+0.6)--(10.3,0+0.6)--(10.3,0-0.6)--(9.2,0-0.6)--cycle;
         \node[text width=3cm] at (12,0) 
         {$\mathbf{\Sigma}$};
         \draw[->,thick](10.3,0)--(11.3,0);
         \node[text width=3cm] at (13.9,0) 
         {$\mathbf{q}$};
         \draw[->,thick](12,0)--(13,0);
          \draw[thick](13.1,0+0.6)--(15.3,0+0.6)--(15.3,0-0.6)--(13.1,0-0.6)--cycle;
           \node[text width=3cm] at (15.7,0) 
         {$f(\mathbf{x}; \mathbf{q})$};
         \draw[->,thick](15.4,0)--(16.4,0);
            \node[text width=3cm] at (19,0) 
         {$\mathbf{y}$};
         
         \draw[dashed] (4.5,-0.5)--(4.5,-\y1+0.5);
    \end{tikzpicture}
    \vspace{-1cm}
    \caption{A single quantized network layer. During training, the weights $\mathbf{w}$ are quantized not only with one, but with $m$ different quantization functions $\mathbf{Q} = \{Q_1, Q_2, \cdots, Q_K\}$, each of them having different number of bits $N=\{n_1,n_2,\cdots,n_K\}$. The resulting quantized weights $\mathbf{q}_k$ are multiplied with attention values $a_k \in [0,1]$ that reflect the importance of the corresponding quantization function $Q_k(\cdot)$. The attention values are optimized during training (cf. Algorithm~\ref{algo:one}). Our learning procedure further applies a temperature scheduling on the attention values that moves from uniform $a_k$ at the beginning of the training to attention values that select only a single quantization function at the end of the training.}
    \label{fig:my_method}
\end{figure*}

In this paper, we introduce \textit{DNN Quantization with Attention} (DQA), a training procedure that can be used to train low-bit quantized DNNs with any quantization method. DQA uses a linear interpolation between quantizers, each at a different precision (c.f. Figure~\ref{fig:my_method}). More specifically, it uses an attention mechanism~\cite{vaswani2017attention} to interpolate, using different importance values for each quantization precision. The importance values are updated during training. Therefore, the DNN has the ability to switch between low, medium and high precision quantization at different stages during training. The importance values involve a temperature term that is progressively cooled down to encourage the attention to focus on one particular quantization precision at the end of the training. We demonstrate, that quantized DNNs trained with DQA consistently outperform quantized DNNs that have been trained with just a single quantization method or with the Binary-Relax scheme, for the same memory and computation budget. In particular, we use DQA to mix uniform min-max quantization of different bitwidths, as well as binary and ternary weight quantization methods. Because DQA improves the performance of existing quantization methods, it is a promising method to deploy DNNs to systems with limited resources. Furthermore, it can be used to apply extreme quantization schemes, such as binarization, to complex tasks or to already compact DNN architectures.

The outline of the paper is as follows. In Section~\ref{sec:related_work} we give an overview of some related works. In Section~\ref{sec:methodology} we introduce the proposed method. Section~\ref{sec:results} presents experiments results and compares the proposed method with other state-of-the-art approaches on challenging computer vision datasets. Finally, we conclude in Section~\ref{sec:conclusion}.

\section{Related Work}
\label{sec:related_work}
Many different methods have been introduced and explored that aim at reducing Deep Neural Networks (DNNs) inference complexity, always with the goal of finding the best trade-off between resource efficiency and model performance. In this section, we introduce some relevant contributions and group them by the type of compression they perform.

\paragraph{Pruning}
Pruning is a compression technique that eliminates several DNN parameters according to a defined criterion in order to reduce its size and complexity. First introduced by~\cite{lecun1990optimal}, pruning received a lot of interest, and numerous contributions have been proposed. For instance, in~\cite{li2016pruning}, the authors use the absolute sum of weights of each channel of a given Convolutional Neural Network (CNN), to select and prune the less important ones. Neuron Importance Score Propagation (NISP)~\cite{yu2018nisp} is another method that estimates the importance of the DNN parameters using the reconstructed error of the last layer before classification when computing back-propagation. Luo et al. define ThiNet~\cite{luo2017thinet}, a pruning method that uses the importance of each feature map in the next layer to prune filters in the current layer. Yamamoto et al.~\cite{yamamoto2018pcas} introduce Pruning Channels with Attention Statistics (PCAS), a pruning method that uses a channel pruning technique based on attention statistics by adding attention blocks to each layer. In the same vein, Shift attention Layer (SAL)~\cite{hacene2019attention} uses an attention mechanism to identify the most important weight in each kernel, prunes the others and replaces a convolution by a shift operation followed by a multiplication. Ramakrishnan et al.~\cite{ramakrishnan2020differentiable} use a learnable mask to identify during learning process the less important parameters, channels or even layers in order to prune. In~\cite{he2020learning}, the authors propose to consider more than one criterion and give the ability to the DNN to decide during training which criterion should be used for each layer when pruning.

\paragraph{Distillation}
Another line of work is distillation that aims at training a quite small DNN called 'student', to reproduce the outputs of another bigger model, called 'teacher'. While initially only considering the final output of the teacher model~\cite{hinton2015distilling}, methods evolved to take into account intermediate representations~\cite{romero2014fitnets, koratana2018lit}. Moreover, some works propose self-distillation~\cite{furlanello2018born} where distilling a model into itself, and show that the student outperforms the teacher while the two networks having the same size and architecture. Other works aim at not only mimicking the outputs of the teacher but also at reproducing the same relations and distances between training examples, yielding a better representation of the latent space for the student, and better generalization capabilities~\cite{park2019relational, lassance2019deep}.

\paragraph{Quantization}

Quantization is another compression technique where a smaller number of bits $n < 32$ is used to represent values. Such an approach reduces the DNN memory footprint since the number of bits required to store its parameters is reduced but also reduces its computational power since operations are computed with a smaller number of bits. Many works have experimentally demonstrated that neural networks do not lose a lot of performance when their parameters are restricted to a small set of possible values~\cite{gupta2015deep}. For instance, in~\cite{choi2018bridging} the authors introduce PArameterized Clipping acTivation (PACT) combined to Statistics-Aware Weight Binning (SAWB), a method that aims at uniformly quantizing both weights and activations on $n$ bits. Learned Step Size Quantization (LSQ)~\cite{esser2019learned} is a quantization method that learns quantization steps in training. Unlike other methods, in backpropagation, it scales the gradient of the scaling factor properly, especially at transition points. Bit-Pruning~\cite{nikolic2020bitpruning} proposes to learn the number of bits each layer requires to represent its parameters and activations during training. In the same vein, Differentiable Quantization of Deep Neural Networks~\cite{uhlich2019differentiable} (DQDNN) tries to combine the features of both LSQ and Bit-Pruning and propose a quantization technique where both number of bits and quantization steps are learned. Other more aggressive quantization methods propose to use low-bit precision up to binarization (resp. ternarization) with only two (resp. three) possible values and one (resp. two) bit storage for each parameter and/or activation~\cite{hubara2016binarized,courbariaux2015binaryconnect,li2016approximate,zhu2016trained,li2016ternary}. Note that reducing precision allows models to be more compact by a great factor, and allows implementation on dedicated low precision hardware~\cite{merolla2014million,farabet2011neuflow,cowan2020automatic,han2020extremely,hacene2018quantized}.

In~\cite{zhou2017incremental}, the authors observed that training quantized networks to low precision benefits from incremental training. Rather than quantizing all the weights at once, they are quantized incrementally by groups with some training iterations between each step. In practice, 50\% of the weights are quantized in the first step, then 75\%, 87.5\% and finally 100\%.  Another method that relates better to our proposed solution is Binary-Relax (BR)~\cite{yin2018binaryrelax}, where a linear interpolation between quantized and full-precision parameters is considered. In BR, a strategy is adopted to push the weights towards the quantized state by gradually increasing the scaling factor corresponding to the quantized parameters. However, such a strategy is handcrafted which may not be the best way to interpolate between quantized and full precision parameters. 

In this contribution, we rely on the fact that the DNN performance can be increased if given the ability to learn other features in addition to its own parameters~\cite{elsken2019neural,ramakrishnan2020differentiable,hacene2019attention,uhlich2019differentiable,nikolic2020bitpruning}, and introduce \textit{DNN Quantization with Attention} (DQA), an attention mechanism-based learnable approach~\cite{vaswani2017attention} where the linear interpolation scaling factors are learned. Moreover, such an approach allows to linearly interpolate with several quantization methods without making it more complex since all scaling factors are learned contrary to BR where we need to find a good way to predetermine them. The learnable approach will converge to the quantization function with the lower number of bits. We demonstrate in this paper that it results in better accuracy for the exact same complexity and number of bits.

\section{Methodology}
\label{sec:methodology}
In this section, we first introduce our learning procedure \textit{DNN Quantization with Attention} (DQA). Later, we review different popular quantization schemes that we use with DQA in our experiments, namely min-max, SAWB, Binary-Weight and Ternary-Weight quantization. In the following, $\mathbf{x}$, $\mathbf{X}$ and $\mathcal{X}$ denotes a vector, a matrix and a set, respectively. 

Let $Q(\mathbf{x}; n)$ be a quantization function that quantizes each element of $\mathbf{x}$ and represents it with $n$ bits.
We consider training of low-bit weight quantized DNNs. In particular, if $f(\mathbf{x}; \mathbf{w})$ is the transfer function of a full precision DNN layer with input $\mathbf{x} \in \mathbb{R}^D$ and weights $\mathbf{w} \in \mathbb{R}^M$, we want to train the corresponding low-bit quantized layer $f(\mathbf{x}; Q(\mathbf{w}; n))$.  Training DNNs with such low-bit quantization can lead to a loss in accuracy compared to the full precision networks due to the reduced capacity of the quantized networks. In~\cite{zhou2017incremental} and later in~\cite{yin2018binaryrelax}, it was observed that low precision DNNs obtain better accuracy when trained incrementally.

Following Figure~\ref{fig:my_method}, let us consider a single quantized network layer with input vector $\mathbf{x}$, 
output vector $\mathbf{y}$ and learnable weights $\mathbf{w}$.
Similar to the idea of Binary-Relax (BR), DAQ relaxes the quantization problem and combines different quantization schemes during training.
More specifically, instead of using just one single $Q(\mathbf{w}; n)$, we propose to train a quantized DNN with a set of $K$ different quantization functions that are averaged during training. More specifically,
\begin{align}
    \mathbf{y} &= f(\mathbf{x}; \mathbf{q}) \\
    \mathbf{q} &= \mathbf{Q}^T \mathbf{a} \\ 
    \mathbf{Q} &= \begin{bmatrix} 
                    Q_1(\mathbf{w}; n_1)^T \\
                    Q_2(\mathbf{w}; n_2)^T \\
                    \vdots \\
                    Q_K(\mathbf{w}; n_K)^T
                    \end{bmatrix},
\end{align}
where $\mathbf{q}$ is the weighted sum of $K$ quantized weight vectors, $\mathbf{Q} \in \mathbb{R}^{K \times M}$ is a matrix whose row vectors are the quantized weight vectors and $\mathbf{a} \in [0,1]^K$ is the attention vector on the quantization functions. Note, that each row of $\mathbf{Q}$ is calculated, using a different quantization function $Q_k(\mathbf{w}; n_k)$ and bitwidth $n_k \in \mathbb{N}$. In particular, we assume that the quantization functions in $\mathbf{Q}$ are sorted by the bitwidth, 
i.e., $n_1 < n_2 < ... < n_K$.

The attention $\mathbf{a}$ is calculated from a soft attention vector $\boldsymbol{\alpha} \in \mathbb{R}^K$, using a softmax function with temperature, i.e.,
\begin{align}
    \mathbf{a} = \frac{e^{\frac{\boldsymbol{\alpha}}{T}}}{\sum_{k=1}^K e^{\frac{\alpha_k}{T}}}, \:\: \in \mathbb{R}^K.
\end{align}
where $T \in \mathbb{R}^+$ is the temperature term. In particular, $\mathbf{a}$ reflects the importance of the $K$ quantization methods $Q_k$. During training, the soft attention $\boldsymbol{\alpha}$ is treated as a trainable parameter that is optimized in parallel to the weights $\mathbf{w}$. In particular, increasing $\alpha_k$ will also increase the corresponding attention weight $a_k$ and therefore the importance of $Q_k(\mathbf{x}; n_k)$. In this manner, the quantized DNN can learn which bitwidth should be used at which stage, during the training.

DQA applies a temperature schedule that cools down the attention $\mathbf{a}$, exponentially
\begin{align}
    T(b) = T(0) \Psi^{b}.
\end{align}
Here, $b = 1, 2, ..., B$ is the batch index for batch-wise training, $T(0) \in \mathbb{R}^+$ is the initial temperature and $\Psi \in [0, 1[$ is the decay rate. Because of that schedule, DQA progressively moves from the full mixture of quantization functions at the beginning of the training to just one single quantization function at the end of training.

In general, training quantized DNNs with such a mixture of different weight quantizations and decaying $T$ will not necessarily result in a quantized DNN that uses a low bitwidth. 
To enforce a low-bit quantized DNN, we therefore augment the loss function with a separate regularizer for each layer
\begin{align}
    r(\boldsymbol{\alpha}) = \frac{\lambda \mathbf{g}^T \mathbf{a}(\boldsymbol{\alpha})}{S} ,
\end{align}
where $S$ is the number of weights in the whole network. Note, that the normalization by $S$ makes the regularizer, and therefore the choice of $\lambda$, independent of the actual network size. $\mathbf{g} = \{g_1,g_2,\cdots,g_K\}$ is a penalty vector, where $g_k$ is increasing with growing $k$. Because we assume, that the quantization functions $Q_k(\mathbf{w}; n_k)$ are sorted by the bitwidth, i.e., $n_1 < n_2 < ... < n_K$ adding $\mathbf{g}^T \mathbf{a}(\boldsymbol{\alpha})$ helps the method to converge to the lowest-bit quantization. Algorithm~\ref{algo:one} summarizes the DQA training.

\begin{algorithm}
\caption{DQA algorithm for a single network layer}
\textbf{Inputs}: Input vector $\mathbf{x}$, initial softmax temperature $T(0)$, final softmax temperature $T(B)$, number of training iterations $B$, and layer transfer function $f$\\
\textbf{Output}: Output tensor $\mathbf{y}$ \\
\begin{algorithmic}
\STATE $\psi = e^{\frac{log \left ( \frac{T(B)}{T(0)} \right ) }{B}} < 1$
\FOR{ each $b=1, 2, ...,B$}
\STATE $T(b) \leftarrow T(0) \psi^b$

\STATE $\boldsymbol{\alpha} \leftarrow {\boldsymbol{\alpha} \over \mathrm{sd}(\boldsymbol{\alpha})}$

\STATE $ \mathbf{a} \leftarrow \mathrm{softmax}(\boldsymbol{\alpha}/T(b))$

\STATE $\mathbf{q} = \mathbf{Q}^T \mathbf{a}$ \text{(linear interpolation)}
\STATE $\mathbf{y} = f(\mathbf{x}, \mathbf{q})$
\STATE  Update $\mathbf{w}$ and $\boldsymbol{\alpha}$ via back-propagation.
\ENDFOR
\end{algorithmic}
\label{algo:one}
\end{algorithm}

In general, DQA is agnostic to the choice of the actual quantization method and can be used with any existing method like min-max, SAWB, binary or ternary quantization. In the following section, we review and define popular quantization methods that we used in our experiments.

\subsection{Choosing the Quantization Functions}
Quantization describes the process of representing a value $x \in \mathcal{X}$ with a corresponding quantized value $q \in \mathcal{Q}$, using a quantization function $Q: \mathcal{X} \rightarrow \mathcal{Q}$. Here, $\mathcal{Q} = \{q_1, q_2, ..., q_{2^n}\}$ is the set of quantization steps that is much smaller than $\mathcal{X}$, i.e., $|\mathcal{Q}| << |\mathcal{X}|$. For a given $w$ and $\mathcal{Q}$, the quantization function minimizes the distance between $w$ and $q$, i.e.,

\begin{align}
    Q(x;n) = \arg\min_{q \in \mathcal{Q}} \|x - q\|, \label{eq:quant_def}
\end{align}
where $\|\cdot\|$ is the Euclidean norm. There are different methods how to construct $\mathcal{Q}$ that yield different quantization schemes, like uniform or non-uniform quantization. 

The first method we may consider is the one introduced in~\cite{nikolic2020bitpruning}. For $\mathcal{X} = [x_{min}, x_{max}]$ they define 
\begin{align}
    q_i = x_{min} + (i-1) \frac{x_{max} - x_{min}}{2^n-1}, \: \: i=1,2,...,2^n.
\end{align}
In particular, the values $q_i$ are uniformly distributed between the values $x_{min}$ and $x_{max}$, what is known as min-max quantization.

The second method we use with our proposed training procedure is Statistics-Aware Weight Binning (SAWB)~\cite{choi2018bridging}. The quantization values are again distributed uniformly over a given interval. However, instead of using the limits $x_{min}$ and $x_{max}$, SAWB introduces a limit $\alpha$, i.e., 
\begin{align}
    q_i = -\alpha + (i-1) \frac{2 \alpha}{2^n-1}, \: \: i=1,2,...,2^n.
    \end{align}
The optimal $\alpha$ can be calculated in a calibration step, using data. In particular, we minimize the mean-square quantization error
\begin{align}
    \alpha^* = \arg\min_{\alpha} \mathrm{E}_{x \sim p(x)}[\|x - Q(x; n, \alpha)\|^2]
\end{align}
with respect to $\alpha$. After calibration, we can use 
$Q(x;n) = Q(x;n,\alpha=\alpha^*)$ for quantization.

For both min-max and SAWB quantization, the solution of
Eq.~\eqref{eq:quant_def} is straight-forward to obtain. It is a uniform quantization function with equally spaced quantization steps that is defined by
\begin{align}
    Q(w;n) = \begin{cases}
                q_1 &, x <= q_1\\
                q_1 + \frac{q_{2^n} - q_1}{2^n - 1} \mathrm{round} \left (x \frac{2^n - 1}{q_{2^n} - q_1} \right ) &, \text{others}\\
                q_{2^n} &, x > q_{2^n}
             \end{cases}.
\end{align}

Another, quantization function worth to mention was introduced for the Binary Weight Network (BWN)~\cite{rastegari2016xnor}. It uses a scaling factor $\beta = \mathbf{E}(|x|)$ and constrains the quantized values to be binary ($n=1$). In particular, with $\mathcal{Q}=\{-\beta, \beta\}$, the quantization function is defined as
\begin{align}
Q(w, 1) = \beta_w \cdot sign(x) = \begin{cases}
            \beta  &,x \geq 0 \\
            -\beta &,\text{others}
    \end{cases}
\end{align}

In the same vein, Ternary Weight Network (TWN)~\cite{li2016ternary} introduces a third quantization step to improve the accuracy. A TWN uses a bitwidth of $n=2$ and a symmetric $\mathcal{Q}=\{-\beta, 0, \beta\}$. Similar to the BWN, the range parameter $\beta$ is calibrated with data. More specifically, we can compute the optimal range $\beta^* = \mathrm{E}_{x~p(x| |x|>\delta)}[|x|]$, where $\delta = 0.7\mathrm{E}[|x|]$ is the symmetric threshold that is used for quantization during calibration.
The resulting quantization function is defined as
\begin{align}
    Q(x; 2) = \begin{cases}
        -\beta &, x \leq - \delta\\
        0 &, |x| \leq \delta \\
        \beta &, x > \delta
    \end{cases}.
\end{align}

\begin{figure*}
    \subfloat[Attention values $a_k$]{\resizebox{0.49\linewidth}{!}{\includegraphics[trim=120 390 80 120,clip]{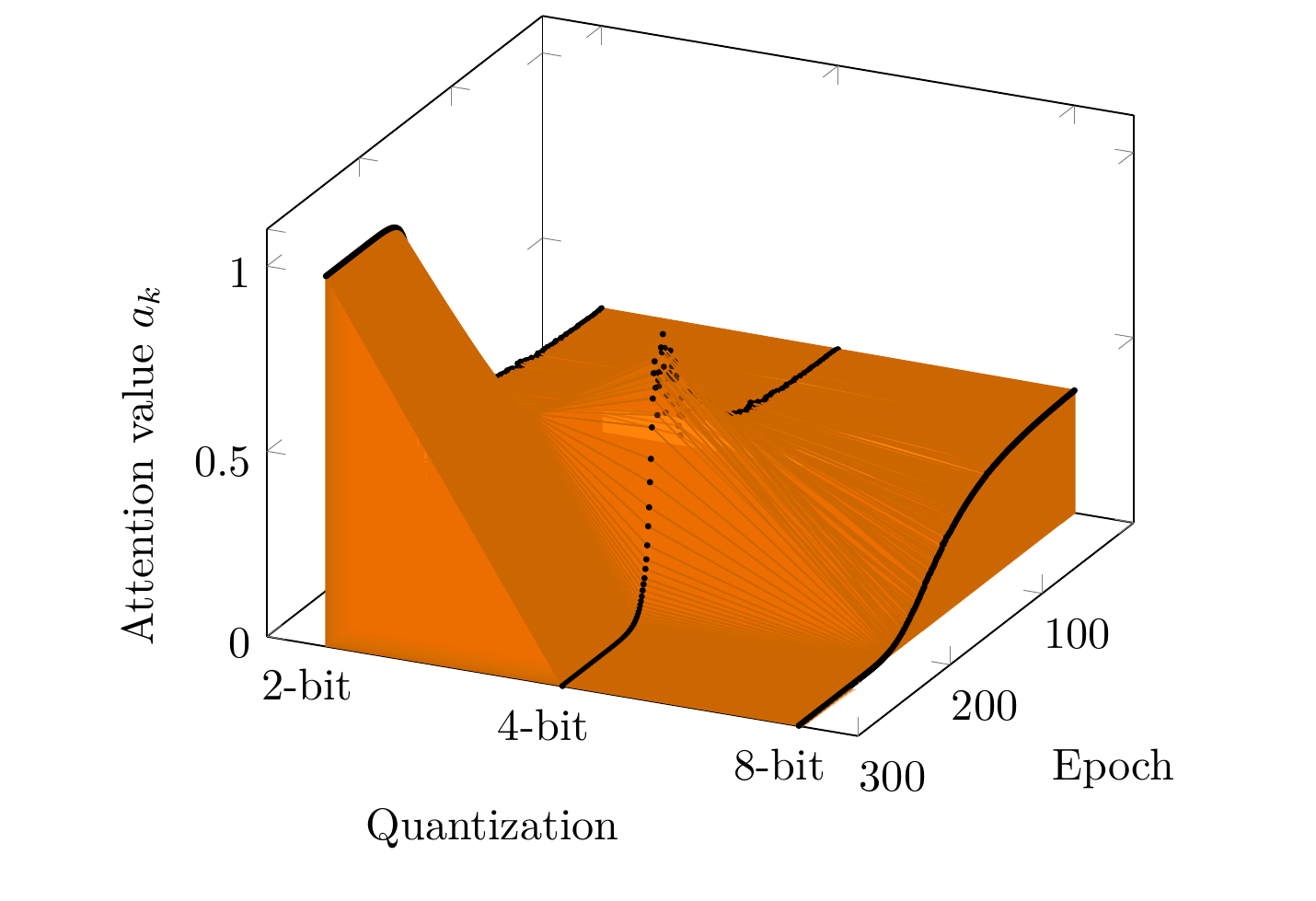}}}
    \hfill
    \subfloat[Quantization function]{\resizebox{0.49\linewidth}{!}{\includegraphics[trim=120 370 80 120,clip]{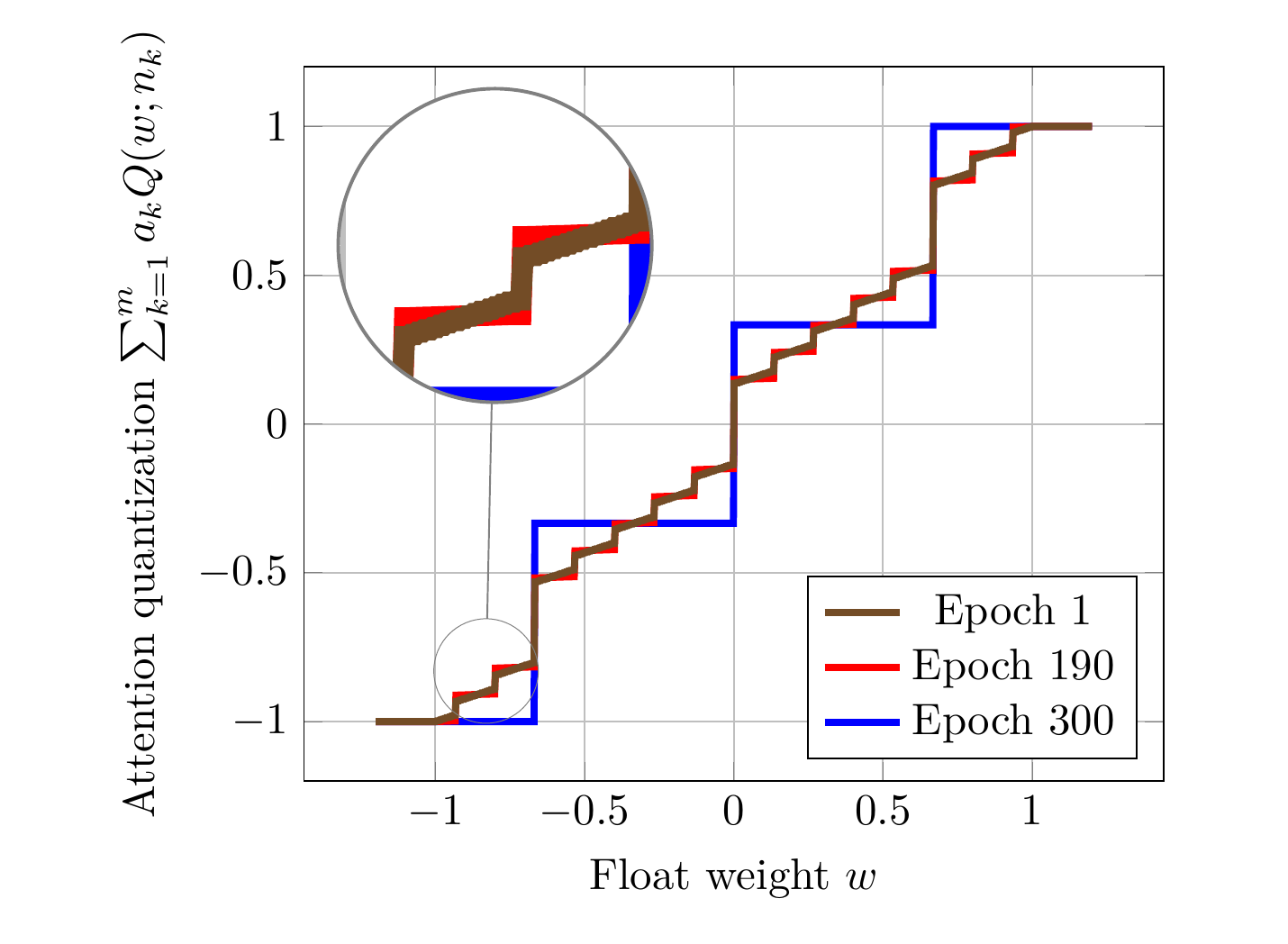}}}
    \caption{Evolution of attention values for proposed DQA training method and corresponding quantization function for the first layer; training done on CIFAR100.}
    \label{fig:evolution_attention}
\end{figure*}

\section{Experiments}
\label{sec:results}
In this section we will first introduce the benchmark protocol used to evaluate our method, then we report different results obtained by DQA and compare them with other counterpart methods.
\subsection{Benchmark Protocol}
To evaluate our method \textit{DNN Quantization with Attention} (DQA), we perform experiments on the three object recognition datasets CIFAR10, CIFAR100 and ImageNet ILSVRC 2012. For each dataset, we use DQA to train low-bit quantized versions of the Resnet18~\cite{he2016deep} and MobileNetV2~\cite{sandler2018mobilenetv2} network architecture. Low-bit means, that we consider networks that only use $n=1$ or $n=2$ Bit for quantization.

For CIFAR10 and CIFAR100, we start from randomly initialized parameters $\mathbf{w}$ and train the quantized networks for $300$ epochs. As an optimizer, we use SGD with an initial learning rate $\gamma =0.1$, which is reduced by a factor of $10$ every $100$ epochs. The training batch size is $128$. 

On the ImageNet ILSVRC 2012 dataset, we train the quantized networks for $90$ epochs, using a batch size of $256$ images. As an initial learning rate, we again use $\gamma=0.1$ which is divided by $10$ every $30$ epochs. That way, we again apply two learning rate drops over the full $90$ epochs. 

For all our experiments, we use DQA with three different quantization functions $\{Q_1, Q_2, Q_3\}$. More specifically, we either consider a mixture of three min-max quantization functions that use $n_1=2$bit, $n_2=4$bit and $n_3=8$bit, respectively or a mixture of BWN, TWN and $8$bit min-max quantization. For the temperature schedule, we use an initial temperature $T(0) = 100$ that is exponentially cooled down to a final value of $T(B) = 0.03$ during training. The soft attention vector is initialized according to
\begin{align}
    \alpha_k = \frac{\sum_{j=1, j\neq k}^N n_j}{\sum_{j=1}^N n_j}.
\end{align}
Note, that since the quantization functions $Q_k(\cdot; n_k)$ are assumed to be sorted by the bitwidth, i.e. $n_1<n_2<\cdots<n_K$, this initialization assigns the highest attention to the quantization function with the lowest bitwidth. The initialization, therefore, acts as a prior that favours low-bit quantized DNNs and therefore helps us to converge to small bit widths early during training. To further encourage low-bit quantized DNNs, we use the penalty values $\mathbf{g}=[1, 4, 16]^T$ that penalizes quantization functions with a large bitwidth.

\subsection{Results}

In the first experiments, we aim at reporting the obtained accuracy achieved by our proposed method, and compare it the baseline full precision network, the baseline quantized network when quantization is performed without any relaxation scheme, and to Binary-Relax (BR) method. To have a fair comparison to BR, we apply BR to the same mixture of quantization functions, i.e.,  
\begin{align}
    \mathbf{q} = \frac{\omega Q_1(\mathbf{w}, n_1) + Q_2(\mathbf{w}, n_2) + Q_3(\mathbf{w}, n_3)}{\omega +2},
\end{align}
where $\omega$ is initialised to $1$ and multiplied by $1.02$ after each epoch.

Table~\ref{tab:cifar10/100} shows the experimental results for the CIFAR10 and CIFAR100 datasets. We report the final validation accuracy of the quantized DNNs for different network architectures and different choices of the quantization functions $\{Q_1, Q_2, Q_3\}$. In general, all reported validation accuracies are the result of a single training run. Only for the experiments that use BWN quantization, we report the average validation accuracy computed over $5$ runs, because the convergence of BWN quantized networks proved to be noisy, which shadowed the effects of DQA. Our proposed method archives comparable accuracy to full precision baseline, and outperforms quantized baseline and BR method when performing BWN, TWN, SAWB and min-max quantization.

The second experiment aims at studying the behavior of the attention values $a_k$ during training. Figure~\ref{fig:evolution_attention} shows the evolution of the attention values $a_k$ and the corresponding quantization function. We can observe from (a) that the attention values have uniform values at start but -- due to the penalty term and the temperature schedule -- slowly converge towards a maximum attention value for the $2$-bit quantization. This evolution can also be seen in (b) where we have a smoother quantization at the start which converges more and more towards the $2$-bit quantization curve. This smooth transition is the reason why DQA yields better results than using a fixed quantization.

The third experiment compares our proposed method with full precision and quantized baselines ImageNet ILSVRC 2012. Table~\ref{tab:resulst_ImageNet} shows that DQA outperforms BR when considering both BWN and min-max quantization. Moreover, DQA reduces significantly the drop in accuracy when quantizing MobileNetV2, and thus may represent a promising lead to apply quantization methods on lightweight DNN architectures.

\begin{table*}
    \centering
    \begin{tabular}{|l|l|l|l|l|l|l|l|l|l|}
    \cline{2-10}
      \multicolumn{1}{l|}{}& Dataset &$n_1$ & $Q_1$ & $n_2$ & $Q_2$ & $n_3$ & $Q_3$  & $\lambda$ & Acc\\
    \hline
     Resnet18    & CIFAR10 & $32$ & FP & - & - & - & - & - & $95.2\%$\\
     \hline
     \hline
     Resnet18    & CIFAR10 & $2$ & min-max & - & - & - & - & - & $91.5\%$\\
     \hline
     Resnet18+BR    & CIFAR10 & $2$ & min-max & $32$ & FP & - & - & - & $93.0\%$\\
     \hline
     Resnet18+BR    & CIFAR10 & $2$ & min-max & $4$ & min-max & $8$ & min-max & - & $93.7\%$\\
     \hline
     Resnet18+Ours    & CIFAR10 & $2$ & min-max & $4$ & min-max & $8$ & min-max & $5$ & $\mathbf{94.8\%}$\\
     \hline
     \hline
     Resnet18    & CIFAR10 & $2$ & SAWB & - & - & - & - & - & $94.8\%$\\
     \hline
     Resnet18+BR    & CIFAR10 & $2$ & SAWB & $4$ & SAWB & $8$ & SAWB & - & $95.1\%$\\
     \hline
     Resnet18+Ours    & CIFAR10 & $2$ & SAWB & $4$ & SAWB & $8$ & SAWB & $1$ & $\mathbf{95.4\%}$\\
     \hline
     \hline
     Resnet18    & CIFAR10 & $1$ & BWN & - & - & - & - & - & $93.8\%$\\
     \hline
     
     Resnet18+BR    & CIFAR10 & $1$ & BWN & $2$ & TWN & $32$ & FP & - & $94.2\%$\\
     \hline
     Resnet18+Ours  & CIFAR10 & $1$ & BWN & $2$ & TWN & $32$ & FP & $5$ & $\mathbf{94.5\%}$\\
     \hline
     \hline
     Resnet18    & CIFAR10 & $2$ & TWN & - & - & - & - & - & $94.3\%$\\
     \hline
     Resnet18+BR    & CIFAR10  & $2$ & TWN & $4$ & min-max & $8$ & min-max & - & $94.5\%$\\
     \hline
     Resnet18+Ours    & CIFAR10  & $2$ & TWN & $4$ & min-max & $8$ & min-max & - & $\mathbf{94.8}\%$\\
     \hline
     \hline
     \hline
      Resnet18    & CIFAR100 & $32$ & FP & - & - & - & - & - & $77.9\%$\\
     \hline
     
     \hline
     Resnet18    & CIFAR100 & $2$ & min-max & - & - & - & - & - & $70.0\%$\\
     \hline
     Resnet18+BR    & CIFAR100 & $2$ & min-max & $32$ & FP & - & - & - & $72.9\%$\\
     \hline
     Resnet18+BR    & CIFAR100 & $2$ & min-max & $4$ & min-max & $8$ & min-max & - & $74.0\%$\\
     \hline
     Resnet18+Ours    & CIFAR100 & $2$ & min-max & $4$ & min-max & $8$ & min-max & $10$ & $\mathbf{76.4\%}$\\
     \hline
     \hline
     Resnet18    & CIFAR100 & $2$ & SAWB & - & - & - & - & - & $77.0\%$\\
     \hline
     Resnet18+BR    & CIFAR100 & $2$ & SAWB & $4$ & SAWB & $8$ & SAWB & - & $77.3\%$\\
     \hline
     Resnet18+Ours    & CIFAR100 & $2$ & SAWB & $4$ & SAWB & $8$ & SAWB & $5$ & $\mathbf{78.1\%}$\\
     \hline
     \hline
     Resnet18    & CIFAR100 & $1$ & BWN & - & - & - & - & - & $75.0\%$\\
     \hline
     Resnet18+BR    & CIFAR100 & $1$ & BWN & $2$ & TWN & $32$ & FP & - & $75.3\%$\\
     \hline
     Resnet18+Ours  & CIFAR100 & $1$ & BWN & $2$ & TWN & $32$ & FP & $30$ & $\mathbf{75.9\%}$\\
     \hline
     \hline
     Resnet18    & CIFAR100 & $2$ & TWN & - & - & - & - & - & $76.1\%$\\
     \hline
     Resnet18+BR    & CIFAR100 & $2$ & TWN & $4$ & min-max & $8$ & min-max & - & $76.3\%$\\
     \hline
     Resnet18+Ours  & CIFAR100 & $2$ & TWN & $4$ & min-max & $8$ & min-max & $20$ & $\mathbf{76.7\%}$\\
     \hline
    \end{tabular}
    \caption{Obtained accuracy of Resnet18 trained on CIFAR10 and CIFAR100, when considering numerous quantization functions (min-max, SAWB, BWN and TWN). Note that FP refers to full precision (i.e. $Q(\mathbf{w},32) = \mathbf{w}$).}
    \label{tab:cifar10/100}
\end{table*}

\begin{table*}
    \centering
    \begin{tabular}{|l|l|l|l|l|l|l|l|l|}
    \cline{2-9}
      \multicolumn{1}{l|}{}&$n_1$ & $Q_1$ & $n_2$ & $Q_2$ & $n_3$ & $Q_3$  & $\lambda$ & Top-1 (Top-5)\\
    \hline
     Resnet18    & $32$ & FP & - & - & - & - & - & $69.9\%$ ($89.1\%$) \\
     \hline
     Resnet18  & $2$ & min-max & - & - & - & - & - & $58.7\%$ ($81.9\%$)\\
     \hline
     Resnet18+Ours & $2$ & min-max & $4$ & min-max & $8$ & min-max & $1$ & $\mathbf{66.9}\%$ ($\mathbf{87.4}\%$)\\
     \hline
     \hline
     MobileNetV2  & $32$ & FP & - & - & - & - & - & $69.0\%$ ($89.0\%$) \\
     \hline
     MobileNetV2 & $2$ & min-max & - & - & - & - & - & $44.2\%$ ($69.8\%$)\\
     \hline
     MobileNetV2+Ours & $2$ & min-max & $4$ & min-max & $8$ & min-max & $1$ & $\mathbf{52.2}\%$ ($\mathbf{77.1}\%$)\\
     \hline
     \hline
     Resnet18 & $1$ & BWN & - & - & - & - & - & $61.0\%$ ($83.5\%$)\\
     \hline
     Resnet18+Ours & $1$ & BWN & $2$ & TWN & $8$ & min-max & $10$ & $\mathbf{61.4}\%$ ($\mathbf{83.7}\%$)\\
     \hline
    \end{tabular}
    \caption{Experiments on the ImageNet dataset, using the Resnet18 and the MobileNetV2 networks. Quantized DNNs trained with DQA consistently outperform quantized DNNs that have been trained with just a single quantization method. It also drastically reduces the accuracy drop when quantizing MobilenetV2.}
    \label{tab:resulst_ImageNet}
\end{table*}


\section{Conclusion}
\label{sec:conclusion}
In this paper, we introduced DQA, a novel learning procedure for training low-bit quantized DNNs. Instead of using only a single quantization precision during training, DQA relaxes the problem and uses a mixture of high, medium and low-bit quantization functions. Our experiments on popular object recognition datasets, such as CIFAR10, CIFAR100 and ImageNet ILSVRC 2012, show that DQA can be used to train highly accurate low-bit quantized DNNs that achieve almost the same accuracy as a full precision DNN with float32 weights. 

Compared to other training procedures that only use a single quantization precision and bitwidth during training, DQA considerably reduces the accuracy drop caused by the quantization. In particular, DQA shows a less significant drop in accuracy when quantizing lightweight DNN architectures such as the MobileNetV2. Such network architectures are already designed to be small and therefore are naturally harder to compress. 

DQA also compares favourably to Binary-Relax (BR), another training procedure for quantized DNNs that applies a mixture of quantized and full-precision weights during training. However, while BR uses a fixed scheme to mix the network weights of different precisions, DQA can learn how to mix them in an optimal way and how to gradually move from high precision to low precision. In practice, this helps training and results in quantized DNNs with higher accuracy.

Most importantly, DQA is agnostic to and can be used with many different existing quantization methods, such as min-max, SAWB, Binary-Weight and Ternary-Weight quantization. Therefore, DQA is a very promising extension to existing DNN quantization methods.

{\small
\bibliographystyle{ieee_fullname}
\bibliography{egbib}
}

\end{document}